\documentclass[a4paper,fleqn]{cas-dc}
\usepackage{balance}
\usepackage{pifont}
\graphicspath{{figure/}}

\usepackage[numbers]{natbib}

\def\tsc#1{\csdef{#1}{\textsc{\lowercase{#1}}\xspace}}
\tsc{WGM}
\tsc{QE}

\begin{document}
\begin{sloppypar}
\let\WriteBookmarks\relax
\def\floatpagepagefraction{1}
\def\textpagefraction{.001}

\shorttitle{AdaFuse: Adaptive Medical Image Fusion Based on Spatial-Frequential Cross Attention}

\shortauthors{Xianming Gu \emph{et al.}}  

\title [mode = title]{AdaFuse: Adaptive Medical Image Fusion Based on Spatial-Frequential Cross Attention}  



%


\author[1]{Xianming Gu}[orcid=0000-0002-1712-702X]



\ead{xianming_gu@foxmail.com}


\credit{Conceptualization, Methodology, Writing-Original Draft}

\author[1]{Lihui Wang}[orcid=0000-0002-3558-5112]
\cormark[1]

\ead{lhwang2@gzu.edu.cn}
\credit{Data Curation, Funding Acquisition, Conceptualization, Reviewing and Editing}

\author[1]{Zeyu Deng}[orcid=0000-0002-4990-441X]
\ead{dengzeyu1996@sina.com}
\credit{Data Curation, Supervision, Investigation, and Resources}

\author[1]{Ying Cao}
\ead{yshxdm@gmail.com}
\credit{Visualization and Validation}

\author[1]{Xingyu Huang}[orcid=0009-0003-0894-8322]
\ead{gs.xingyuhuang21@gzu.edu.cn}
\credit{Review \& Editing}

\author[2]{Yue-min ZHU}
\ead{yue-min.zhu@insa-lyon.fr}
\credit{Review \& Editing. All the authors reviewed the manuscript}

%
\affiliation[1]{organization={Engineering Research Center of Text Computing \& Cognitive Intelligence, Ministry of Education, Key Laboratory of Intelligent Medical Image Analysis and Precise Diagnosis of Guizhou Province, State Key Laboratory of Public Big Data, College of Computer Science and Technology, Guizhou University, Guiyang 550025, China}
}
        
\affiliation[2]{organization={University Lyon, INSA Lyon, CNRS, Inserm, IRP Metislab CREATIS UMR5220, U1206, Lyon 69621, France}
}

\cortext[1]{Corresponding author}



\begin{abstract}
Multi-modal medical image fusion is essential for the precise clinical diagnosis and surgical navigation since it can merge the complementary information in multi-modalities into a single image. The quality of the fused image depends on the extracted single modality features as well as the fusion rules for multi-modal information. Existing deep learning-based fusion methods can fully exploit the semantic features of each modality, they cannot distinguish the effective low and high frequency information of each modality and fuse them adaptively. To address this issue, we propose AdaFuse, in which multi-modal image information is fused adaptively through frequency-guided attention mechanism based on Fourier transform. Specifically, we propose the cross-attention fusion (CAF) block, which adaptively fuses features of two modalities in the spatial and frequency domains by exchanging key and query values, and then calculates the cross-attention scores between the spatial and frequency features to further guide the spatial-frequential information fusion. The CAF block enhances the high-frequency features of the different modalities so that the details in the fused images can be retained. Moreover, we design a novel loss function composed of structure loss and content loss to preserve both low and high frequency information. Extensive comparison experiments on several datasets demonstrate that the proposed method outperforms state-of-the-art methods in terms of both visual quality and quantitative metrics. The ablation experiments also validate the effectiveness of the proposed loss and fusion strategy.
\end{abstract}



\begin{keywords}
 \sep Medical image fusion \sep Spatial-frequential cross attention \sep Transformer \sep Fourier transform
\end{keywords}

\maketitle

\section{Introduction}
\label{sec:introduction}
\begin{figure}[!t]
	\centerline{\includegraphics[width=\linewidth]{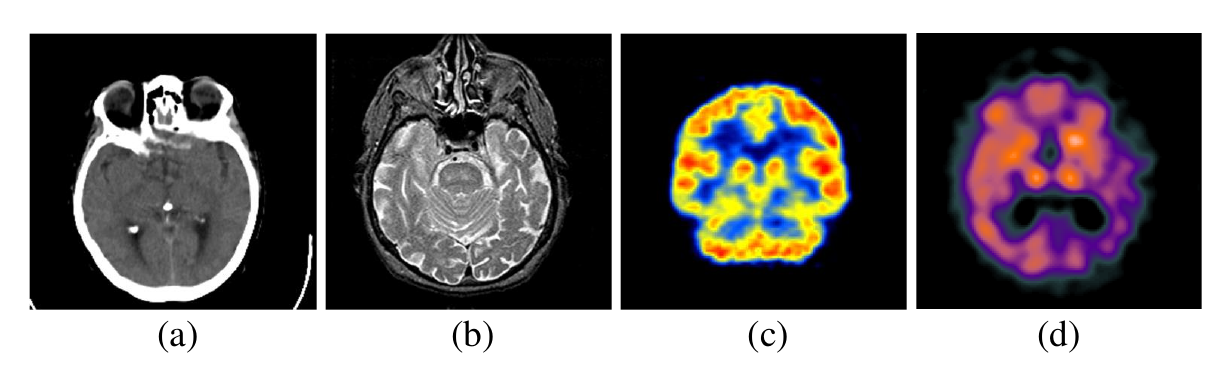}}
	\caption{Examples of medical images from different modalities. (a)CT images can provide bone density information; (b)MRI images can provide high-contrast and high-resolution soft tissue structures; (c)PET and (d)SPECT can reflect physiological activities such as metabolism, molecular motion, and blood flow within the human body.}
	\label{fig_med_img}
\end{figure}

Multi-modal medical imaging provides more comprehensive complementary information for clinical diagnosis. Multi-modal medical images obtained from different sensors represent different physiological information of the human body, as shown in Fig. \ref{fig_med_img}. Therefore, effective fusion of information from multi-modal medical images is of crucial importance in enhancing clinical diagnosis \cite{shahdoosti2018multimodal,prakash2019multiscale}, personalized treatment, disease monitoring and prognosis prediction \cite{rajalingam2018review}.

Existing medical image fusion methods can be categorized into traditional methods and deep learning-based methods. In traditional fusion methods, the extraction and fusion of multi-modal image features are typically performed in either the spatial or transform domain \cite{azam2022review}. In the transform domain, source images are decomposed into frequency domain feature coefficients or frequency sub-bands using various transform techniques. Fusion rules are then applied to merge these coefficients, followed by inverse transforms to reconstruct the fused image. Examples of such methods include complex wavelet transform \cite{singh2014fusion}, Laplacian pyramid \cite{du2016union}, discrete cosine transform \cite{jain2021multimodal}. Li \emph{et al.} \cite{li2020laplacian} introduced the Laplacian recombination method, which leverages the Laplacian decision map to capture complementary, redundant, and low-frequency information. High-frequency sub-bands of the fused image are reconstructed by utilizing the global decision map and local mean, followed by the Laplacian inverse transform to obtain the final fusion result. Yin \emph{et al.} \cite{Yin2019NSSTPAPCNN} combined the nonsubsampled shearlet transform (NSST) with the parameter adaptive pulse coupled neural network (PA-PCNN) for medical image fusion. They involve decomposing the source images into NSST sub-bands, and extracting and fusing high-frequency and low-frequency coefficients using the PA-PCNN model, and finally reconstructing the fused image through inverse NSST. In the spatial domain, fusion methods rely on saliency measurements to select regions or pixels from the source images and apply linear or nonlinear transform operations for fusion \cite{wang2021salient}. Sparse representation \cite{zhang2018sparse} and dictionary learning \cite{zhu2016novel} are examples of techniques employed in this domain. Liu \emph{et al.} \cite{liu2019csmca} proposed CSMCA, which combines convolutional sparse representation with morphological component analysis for CT-MRI image fusion. However, due to the large difference in saliency measures of multi-modal images, fusion results may suffer from distortions \cite{el2016current}. Both spatial and transform domain fusion methods often require manual design of feature extraction techniques and fusion rules, leading to the dependence of fusion image quality on the rationality of these designs and lacking robustness and adaptiveness. Additionally, many traditional methods suffer from high computational complexity and long execution times.

In recent years, deep learning (DL) has gained increasing attention due to its powerful feature extraction capabilities \cite{salau2019feature}. DL-based fusion methods usually use neural networks for feature extraction and fusion, avoiding the pitfalls of manual design. Liu \emph{et al.} \cite{Liu2017MedCNN} applied convolutional neural networks (CNNs) to medical image fusion, using features extracted by CNN to calculate fusion weights and combining with Laplace pyramid for feature fusion. Zhang \emph{et al.} \cite{ZHANG2020IFCNN} proposed a unified supervised fusion method called IFCNN, which utilizes a pre-trained ResNet101 \cite{he2016deep} as the feature extraction framework. However, this method still relies on manually designed rules for feature fusion and requires a large amount of labeled data for model training. Ma \emph{et al.} \cite{Ma2020DDcGAN} proposed a multi-resolution image fusion method called DDcGAN, which utilizes convolutional and deconvolutional layers for feature extraction and employs two discriminators to constrain the similarity between the fused result and the two source images. However, this method suffers from training difficulties, slow convergence, and may result in poor fusion quality. Zhang \emph{et al.} \cite{zhang2020PMGI} proposed a multi-scale fusion framework called PMGI, which maintains proportion gradient and intensity information to avoid information loss caused by convolutional layers. Xu \emph{et al.} \cite{Xu2020U2Fusion} presented a unified unsupervised image fusion network called U2Fusion, which employs DenseNet dense convolutional blocks and calculates fusion weights adaptively based on the source images. Another approach by Zhang \emph{et al.} \cite{zhang2021sdnet} designed a compression-reconstruction fusion network called SDNet, which models fusion as an image compression-reconstruction task using convolutional layers for feature extraction and reconstruction, with constraints to maintain similarity with the source images. However, this method may lead to the loss of crucial features from each modality. Recently, Han \emph{et al.} \cite{han2023ie} inspired by the information exchange mechanism \cite{nie2020multi}, proposed IE-CFRN for multi-modal medical image fusion, using group convolution to construct a channel-wise information exchange network and the important degree of relation between two source images can be learned automatically. Ding \emph{et al.} \cite{ding2023m4fnet} proposed M$^4$FNet, which used multi-scale functions combined with dilation convolution to increase multi-scale-receptive field, and preserved the long-range relationships in deep feature blocks. While CNN-based image fusion has made significant progress, extracting features using uniform convolutional kernels from different modality images may not be the most effective method and can result in the loss of detailed information in the fused image. Moreover, CNNs have limited receptive fields and mainly focus on local features, neglecting global context and long-range dependencies. 

With the successful application of Transformer \cite{vaswani2017attention} in computer vision \cite{dosovitskiy2020vit,liu2021swin}, VS \emph{et al.} \cite{vs2022ift} proposed the Image Fusion Transformer (IFT), which uses multi-head self-attention mechanisms to design an novel feature extraction strategy for multi-modal image fusion. Ma \emph{et al.} \cite{Ma2022SwinFusion} combined the Swin Transformer \cite{liu2021swin} and introduced a general-purpose image fusion Transformer termed as SwinFusion. It utilizes a cross-domain remote learning module to fuse features from different modalities and achieves promising results. Zhang \emph{et al.} \cite{zhang2023multimodal} proposed a self-supervised transformer for multi-modal image fusion termed as SSTFusion, designed random image super-resolution as the pretext task to train the network. However, their method may loss the important information of medical images. Although Transformers have dominated the field of image processing in recent years, they have limited capability in extracting high-frequency features and may overlook crucial high-frequency information \cite{park2022vision}. Furthermore, Transformers often have a large number of parameters and require large-scale data for training, which is challenging to obtain in the medical imaging domain.
In summary, most deep learning-based fusion methods utilize CNNs for feature extraction, which are effective in capturing local features but overlook the global context. Although Transformer \cite{vaswani2017attention} can model global features, they have limitations in extracting high-frequency features and may result in the loss of crucial texture details. Furthermore, many deep learning-based methods still employ strategies such as maximum selection \cite{li2020laplacian} or L1-norm fusion rules \cite{li2018densefuse} to fuse features, lacking adaptability. Thirdly, existing deep learning-based fusion methods often perform feature extraction and fusion at a single scale, neglecting the rich multi-scale feature information in the source images, which can lead to information loss. Lastly, existing deep learning-based fusion methods often focus on modeling a single type of medical image pair (e.g., CT-MRI or PET-MRI) and do not consider the generalizability of the model to different types of medical images.

To address the limitations, this paper proposes an unsupervised end-to-end adaptive unified fusion model termed as AdaFuse for multi-modal medical image fusion. Firstly, we use convolutional layers and max-pooling operations to extract multi-scale local features from the source images, and then apply Transformer to extract global contextual information. To compensate for the loss of high-frequency information in the Transformer model, we introduce a spatial-frequency domain feature combination approach. We incorporate Fourier Transform-guided fusion branches to better preserve high-frequency features and global characteristics in the fusion results. Additionally, to adaptively fuse the extracted multi-modal features, we design a Cross Attention Fusion (CAF) block that learns the weights of the source images in the fusion results adaptively, without the need for complex manual design. Considering the characteristics of spatial and frequency domain features, we combine structural tensors \cite{di1986note} and design a novel loss function that is better suited for our multi-modal medical image fusion task.

The main contributions of this work are as follows:
\begin{itemize}
	\item We propose a novel unsupervised end-to-end adaptive fusion model called AdaFuse, which enables adaptive feature extraction and fusion without the need for complex manual rules, and then apply AdaFuse for multi-modal medical image fusion.
	
	\item We propose a spatial-frequency domain feature fusion method that incorporates Fourier Transform-guided fusion branches. This approach addresses the limitations of Transformer in capturing high-frequency information, thereby enabling better preservation of both high-frequency and global features in the fused result.
	
	\item We propose spatial-frequential fusion (SFF) module, which consists of cross attention fusion (CAF) block that adaptively fuses features from different modalities of the source image using a multi-headed self-attention mechanism. Such SFF module can effectively preserve important features and mitigate the problem of information loss in the fused images.
	
	\item We extensively evaluate proposed AdaFuse on three medical image fusion datasets: CT-MRI, PET-MRI, and SPECT-MRI. We compare AdaFuse with state-of-the-art models using both qualitative and quantitative evaluation methods. The results demonstrate that our proposed AdaFuse achieves superior performance and generalization capability.
\end{itemize}

\section{Method}

\subsection{Network Architecture}
\begin{figure*}[!t]
	\centerline{\includegraphics[width=0.7\textwidth]{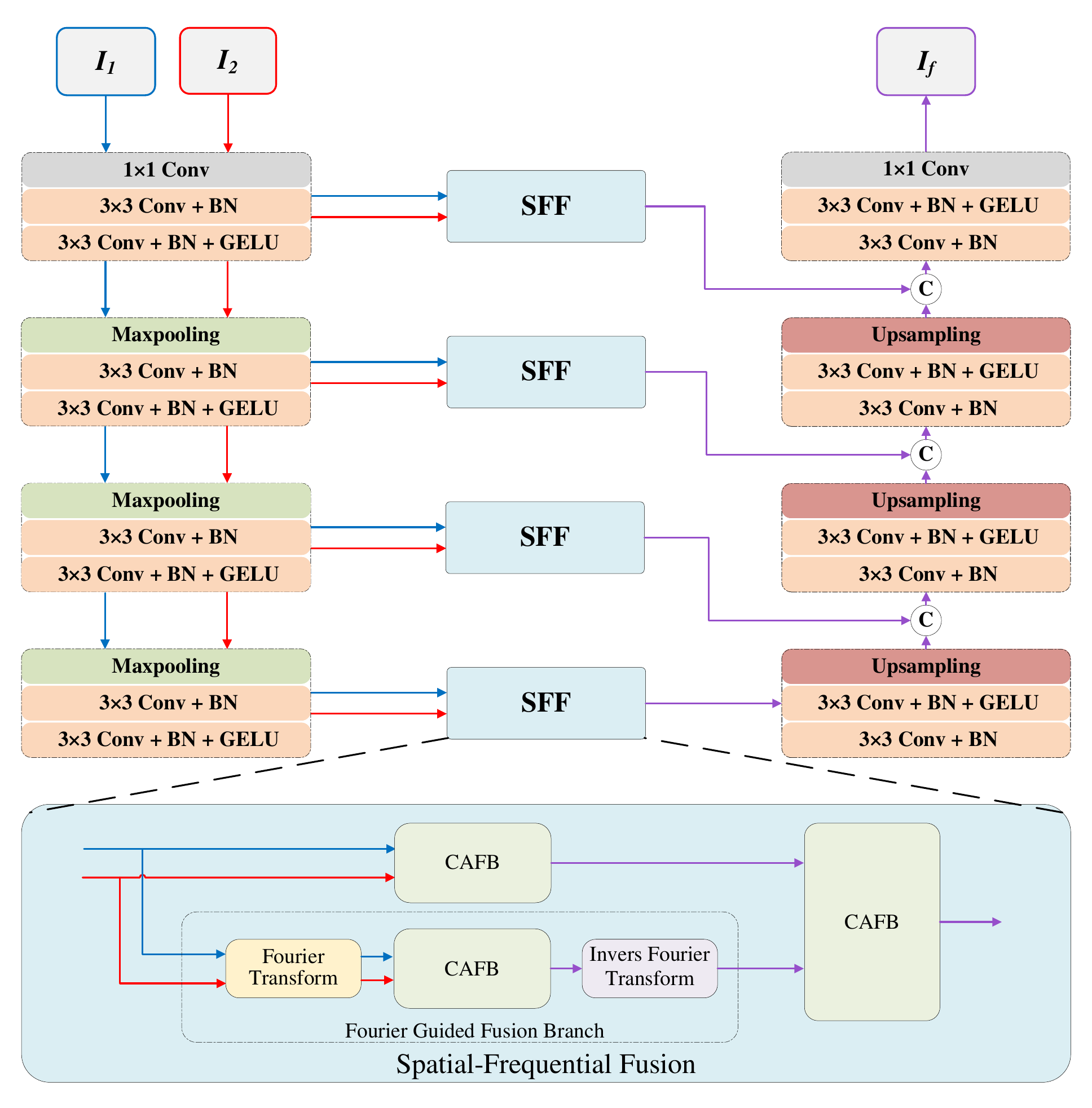}}
	\caption{Architecture of the proposed AdaFuse. Our method performs feature fusion at each of the four scales using a spatial-frequential fusion (SFF) module. In SFF, we designed the Fourier-guided fusion branch (FGFB) to extract the frequency domain features and proposed cross attention fusion (CAF) for cross-domain feature fusion.}
	\label{fig_Framework}
\end{figure*}

The proposed AdaFuse is an end-to-end medical image fusion network, the architecture of which is based on encoder-decoder network, as shown in Fig. \ref{fig_Framework}. AdaFuse takes in a pair of multi-modal input source images ${I_1} \in {\mathbb{R}^{H \times W \times {C_1}}}$ and ${I_2} \in {\mathbb{R}^{H \times W \times {C_2}}}$, generates a complete fused image ${I_f} \in {\mathbb{R}^{H \times W \times {C_f}}}$. $H$ and $W$ denote the height and width of the input and output images. $C_1$, $C_2$, and $C_f$ are the number of channels of the two input images and the fused image, respectively. 
First, we get the multi-scale shallow features $\phi _j^i \in {\mathbb{R}^{{H_j} \times {W_j} \times {C_j}}}$  of input images extracted by the encoder network, with $i \in 1,2$ indicating the index of input images, $j \in \{ 1,2,3,4\} $ indicating four different scales, which is defined as
\begin{equation}
	\phi _j^i =\begin{cases}
		\mbox{Conv}({I_i}),& j=1; \\ 
		\mbox{Conv}(\mbox{MP}(\phi _{j - 1}^i)),& j=2,3,4.
	\end{cases}
	,
\end{equation}
where ${\text{Conv}}( \cdot )$ represents convolution operations, and ${\text{MP}}( \cdot )$ represents the max-pooling operations. In order to fuse the valid information of ${I_1}$ and ${I_2}$ adaptively, and avoid the lack of high-frequency information, we propose SFF module to obtain the fusion features $\phi _j^f$ from different scales features $\phi _j^i$. As shown in Fig. \ref{fig_Framework}, SFF module contains three CAF blocks, which are used to obtain fusion feature maps $\psi _j^{{f_s}}$ of spatial domain, is defined as
\begin{equation}
	\psi _j^{f_s}=\mbox{CAF}(\phi _j^1,\phi _j^2).
\end{equation}

Considering that the frequency domain can provide supplementary information for spatial domain fusion, we propose a Fourier-guided fusion branch, which uses CAF module to fuse the frequency domain feature maps of images from different sources. To be specific, we first transform the image features of different sources with Fourier transform to obtain the corresponding frequency domain features $\tilde \phi _j^i$, then the frequency domain characteristics of adaptive fusion $\tilde \psi _j^{{f_f}}$ are obtained by using CAF module for fusion. Finally, after the inverse Fourier transform, we can obtain the adaptive fusion features $\psi _j^f$ based on frequency information $\psi _j^{{f_f}}$. In view of the fact that frequential fusion features and spatial fusion features can complement and guide each other, the cross attention of frequential domain fusion features and spatial domain fusion features are calculated respectively, and the adaptive fusion is carried out to obtain the fusion features $\psi _j^f$. The overall process can be formulated as
\begin{equation}
	\begin{aligned}
		&\tilde{\phi} _j^i=\log(\left|\mbox{FT}(\phi _j^i)\right|+\epsilon), \\ 
		&\tilde \psi _j^{f_f}=\mbox{CAF}(\tilde \phi _j^1,\tilde \phi _j^2), \\
		&\psi _j^{f_f}=\mbox{IFT}(\tilde \psi _j^{f_f}), \\
		&\psi _j^{f}=\mbox{CAF}(\psi _j^{f_s},\psi _j^{f_f}).
	\end{aligned}
	,
\end{equation}
where ${\text{FT}}( \cdot )$ represents the Fourier transform and ${\text{IFT}}( \cdot )$ denotes inverse Fourier transform. $\epsilon$ ensures that the calculation makes sense.

Finally, we use the decoder network (the right part of Fig. \ref{fig_Framework}) to reconstruct the multi-scale fusion features $\psi _j^f$ , and obtain the fused image ${I_f}$, which can be formulated as
\begin{equation}
	\label{Eq4}
	\begin{split}
	I_f=&\mbox{Conv}([\uparrow \mbox{Conv}([\uparrow \mbox{Conv}([\\&\uparrow \mbox{Conv}(\psi _4^{f}),\psi _3^{f}]),\psi _2^{f}]),\psi _1^{f}]),
	\end{split}
\end{equation}
where $\uparrow$ denotes up-sampling operation.

\begin{figure*}[!t]
	\centerline{\includegraphics[width=\textwidth]{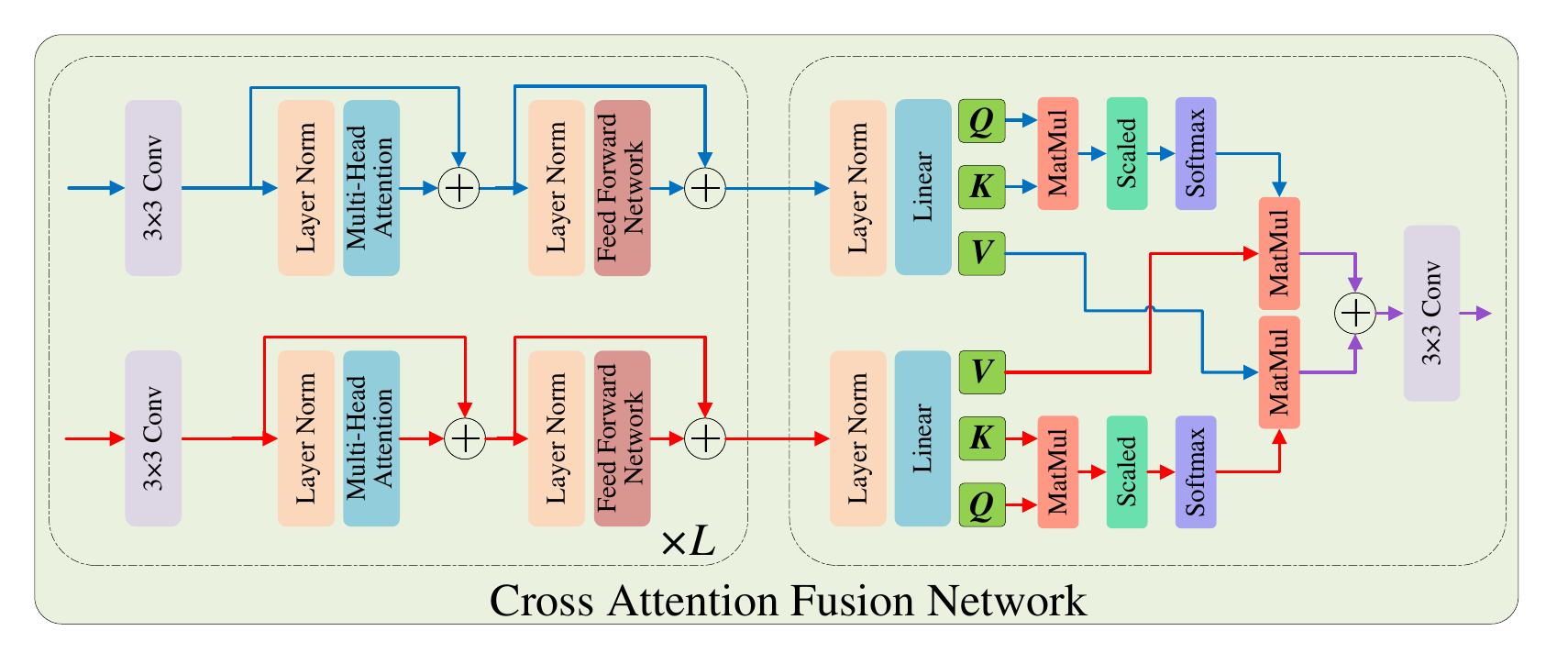}}
	\caption{The Details of cross attention fusion (CAF) Block. Features are fused by crossing the attention scores of different modes}
	\label{fig_FusionNetwork}
\end{figure*}

In order to obtain high-quality fusion image, we need to adaptively fuse the salient features of different input images to obtain effective fusion feature representation. For this purpose, we propose CAF, which is shown in Fig. \ref{fig_FusionNetwork}.

Specifically, for the two features ${\phi ^1}$ and ${\phi ^2}$ that need to be fused, we first use Transformer Encoder structure to obtain global deep features $\phi _{global}^1$ and $\phi _{global}^2$ of different inputs, defined as
\begin{equation}
	\begin{aligned}
		&F^{i}=\mbox{PE}(\phi^i)+\mbox{MSA}(\mbox{LN}(\mbox{PE}(\phi^i))) \\
		&\phi ^{i}_{global}=\phi ^i+\mbox{FFN}(\mbox{LN}(F^i))
	\end{aligned}
	\quad \mbox{with } i=1,2,
\end{equation}
where ${\text{PE}}( \cdot )$ represents the patch embedding operation. Suppose the original input feature map has a size of $h \times w \times c$ and the patch size is $p \times p$, the feature map size becomes $\frac{h}{p} \times \frac{w}{p} \times ({p^2} \times c)$ after the PE operation. ${\text{LN}}( \cdot )$ denotes Layer Normalization, ${\text{MSA}}( \cdot )$ represents the Multi-head Self-Attention operation, and ${\text{FFN}}( \cdot )$ is the Feed-forward network composed of two fully connected layers and the Gaussian Error Linear Unit (GELU) activation function.

After obtaining the global feature representations $\phi _i^{global} \in {\mathbb{R}^{\frac{h}{p} \times \frac{w}{p} \times ({p^2} \times c)}}$ for different inputs, in order to adaptively fuse the features from different inputs, we introduce self-attention mechanism. Specifically, we apply layer normalization to the global features, and then use three learnable $d$-dimension weight matrices ${{\mathbf{W}}^Q}$, ${{\mathbf{W}}^K}$, ${{\mathbf{W}}^V} \in {\mathbb{R}^{({p^2} \times c) \times d}}$ to map them into ${{\mathbf{Q}}^i}$, ${{\mathbf{k}}^i}$, and ${{\mathbf{V}}^i} \in {\mathbb{R}^{(\frac{h}{p} \times \frac{w}{p}) \times d}}$ vectors respectively, which is defined as
\begin{equation}
	\begin{array}{*{20}{c}}
		{{{\mathbf{Q}}^i}{\text{  =  LN}}(\psi _{global}^i){{\mathbf{W}}^Q}} \\ 
		{{{\mathbf{K}}^i}{\text{  =  LN}}(\psi _{global}^i){{\mathbf{W}}^K}} \\ 
		{{{\mathbf{V}}^i}{\text{  =  LN}}(\psi _{global}^i){{\mathbf{W}}^V}} 
	\end{array}{\text{ }},{\text{with }}i = 1,2.
\end{equation}

To perform feature fusion adaptively, we propose cross attention, by exchanging the key vectors between the two modalities, we can compute the similarity between the two modalities and then use the similarity scores ${s^i}$ for adaptive fusion. Mathematically, it can be defined as
\begin{equation}
	\begin{aligned}
		s^1=\mbox{softmax}(\frac{\mathbf{Q}^1{\mathbf{K}^2}^T}{\sqrt{d}}) \\
		s^2=\mbox{softmax}(\frac{\mathbf{Q}^2{\mathbf{K}^1}^T}{\sqrt{d}})
	\end{aligned}
	\quad,
\end{equation}
where larger value of ${s^i}$ indicates higher similarity to another modality. Then we can determine the fused feature representation through the CAF. The fused feature is obtained as follows:
\begin{equation}
	\begin{aligned}
		\psi^{f}&=\mbox{CAF}(\phi^1,\phi^2)\\
		&=(1-s^1)\mathbf{V^1} \oplus  s^1\mathbf{V^2}\oplus (1-s^2)\mathbf{V^2} \oplus  s^2\mathbf{V^1}.
	\end{aligned}
\end{equation}
where $ \oplus $  represents the element-wise summation operation. Then we can obtain the fused features $\psi _1^f$, $\psi _2^f$, $\psi _3^f$, and $\psi _4^f$ at each scale after SFF. Finally, the fusion image ${I_f}$ can be obtained using \eqref{Eq4}.

\subsection{Loss Function}
To better preserve the detailed information (high-frequency features) and global contextual information (low-frequency features) from the source images, we propose a novel loss function that combines the characteristics of the Fourier transform. This loss function consists of a pixel-wise content loss ${\mathcal{L}_{content}}$ and a structural loss  ${\mathcal{L}_{structure}}$. The pixel-wise content loss ensures that the fused image contains the crucial content from the source images without losing important information. It is defined as
\begin{equation}
	\mathcal{L}_{content}=\sum_{x}^{H}\sum_{y}^{W}\Vert I_f - \frac{I_1+I_2}{2}\Vert_2,
\end{equation}
where $H$  and $W$  represent the height and width of the source images and  $\Vert \cdot \Vert_2$ denotes the ${L_2}$  norm.

The structural loss function consists of logarithmic structural tensor loss and structural similarity (SSIM) loss. The structural tensor ${\mathbf{Z}}_{\mathbf{I}}^{x,y}$ \cite{di1986note}, first used by Jung \emph{et al.} \cite{jung2020unsupervised} for image fusion , can be used to measure the gradient information in all directions of multi-modal images., which is defined as
\begin{equation}
	\mathbf{Z}_{\mathbf{I}}^{x,y} = \left[ {\begin{array}{*{10}{c}}
			{\sum\limits_{i = 1}^M {{{\left( {{\nabla _x}I_i^{x,y}} \right)}^2}} }&{\sum\limits_{i = 1}^M {\left( {{\nabla _x}I_i^{x,y}} \right)\left( {{\nabla _y}I_i^{x,y}} \right)} } \\ 
			{\sum\limits_{i = 1}^M {\left( {{\nabla _y}I_i^{x,y}} \right)\left( {{\nabla _x}I_i^{x,y}} \right)} }&{\sum\limits_{i = 1}^M {{{\left( {{\nabla _y}I_i^{x,y}} \right)}^2}} } 
	\end{array}} \right],
\end{equation}
where  ${I_i}$ represents the $i$-th channel of the $M$ channels image $I$. ${\nabla _x}$ and ${\nabla _y}$ represent the horizontal and vertical gradient operators, respectively. In our method, to further preserve the gradient features from both source images ${I_1}$ and ${I_2}$, we concatenate them along the channel dimension: 
\begin{equation}
	\mathbf{I}_c=\mbox{Concat}(I_1,I_2).
\end{equation}

To improve the effectiveness of frequency guidance, we design a logarithmic loss function based on the structural tensor, which is defined as
\begin{equation}
	{\mathcal{L}_{grad}}=\log(1+\sum_{x}^{H}\sum_{y}^{W} \Vert \mathbf{Z}_{I_f}^{x,y}-\mathbf{Z}_{\mathbf{I}_c}^{x,y}\Vert _F^2).
\end{equation}
where $\Vert \cdot \Vert_F$ denotes the Frobenius norm.

To further ensure that the fused result contains crucial features in terms of luminance, contrast, and structure, we incorporate an SSIM constraint into the structural loss function, which can be formulated as
\begin{equation}
	\mathcal{L}_{ssim} = w_1 \cdot (1-SSIM(I_f,I_1)) + w_2 \cdot (1-SSIM(I_f,I_2)),
\end{equation}
where  $SSIM( \cdot )$ represents the structural similarity operation. ${w_1}$ and ${w_2}$ are weights that control the importance of preserving information from the two source images.

The total structural loss function is formulated as
\begin{equation}
	\mathcal{L}_{structure} =  {\mathcal{L}_{grad}} +  {\mathcal{L}_{ssim}}.
\end{equation}

Finally, the total loss function of our AdaFuse is defined as
\begin{equation}
	\mathcal{L} = \lambda {\mathcal{L}_{content}} +  {\mathcal{L}_{structure}},
\end{equation}
where $\lambda$ is employed to control the trade-off between different losses.

\section{Experiments}
\subsection{Datasets and Experimental Settings}
The datasets used in this study are obtained from the publicly available Harvard medical AANLIB dataset\footnote{\href{http://www.med.harvard.edu/AANLIB/home.html}{http://www.med.harvard.edu/AANLIB/home.html}} . We selected 184 pairs of CT-MRI, 269 pairs of PET-MRI and 357 pairs of SPECT-MRI images from the dataset. The image size is 256×256. For each type of dataset, 24 pairs of images are selected as the testing set, and the remaining pairs are used as the training set. Since PET and SPECT images are color images composed of RGB channels, while CT and MRI images are single-channel grayscale images, for the PET-MRI and SPECT-MRI image fusion tasks, we first convert the PET and SPECT images to the YCbCr color space. The Y (luminance) channel contains the structural and intensity information of the image, while the Cb and Cr (chrominance) channels mainly contain color information. Therefore, during the training of the fusion model, we only use the Y channel to represent the input of SPECT and PET images. The fused image is then mapped back to the RGB color space along with the Cb and Cr channels to obtain the final fusion result.

To evaluate our proposed AdaFuse, we compare it with eleven state-of-the-art multi-modal medical image fusion methods, including CSMCA \cite{liu2019csmca}, MedCNN \cite{Liu2017MedCNN}, IFCNN \cite{ZHANG2020IFCNN}, DDcGAN \cite{Ma2020DDcGAN}, PMGI \cite{zhang2020PMGI}, U2Fusion\cite{Xu2020U2Fusion}, SDNet \cite{zhang2021sdnet}, SwinFusion \cite{Ma2022SwinFusion}, M$^4$FNet \cite{ding2023m4fnet}, IE-CFRN \cite{han2023ie} and SSTFusion \cite{zhang2023multimodal}.

All experiments in this paper were conducted using the PyTorch 1.10 framework on an NVIDIA Tesla A100 GPU. During the training process, we used the AdamW \cite{loshchilov2017adamw} optimizer with a learning rate of $2 \times {10^{ - 4}}$ and trained the model until convergence. The number of epochs was set to 2000, and the batch size was set to 4. In our proposed loss function, the hyperparameters are set to ${w_1} = {w_2} = 0.5$, $\lambda = 0.5$. In the CAF, the patch size $P$ was set to 16, and the embedding dimension was set to 256. Additionally, for CT-MRI structure-structure image fusion, we adopted a shared parameter strategy for the convolutional layers used in feature extraction to reduce the number of parameters. The hyperparameters of the baseline models are kept at their default settings.

\subsection{Objective Evaluation Metrics}
To quantitatively evaluate the fusion performance of different models, we employ five quantitative metrics for comparative analysis: entropy (EN) \cite{roberts2008en}, peak signal-to-noise ratio (PSNR), mutual information (MI) \cite{qu2002mi}, correlation coefficient (CC), and feature mutual information (FMI) based on discrete cosine transform \cite{haghighat2014fmi}. Entropy measures the amount of information contained in the fused image from an information theory perspective. PSNR reflects the distortion level by comparing the peak power to the noise power ratio. MI measures the similarity between two images, indicating how well the fused image preserves the information from the source images. CC measures the linear correlation between the source and fused images. FMI, based on discrete cosine transform feature mutual information, measures the fusion quality. Higher values of these evaluation metrics indicate better fusion results with richer information content.

\subsection{Experimental Results}
\subsubsection{CT-MRI Image Fusion}

\begin{figure*}[!t]
	\centerline{\includegraphics[width=\textwidth]{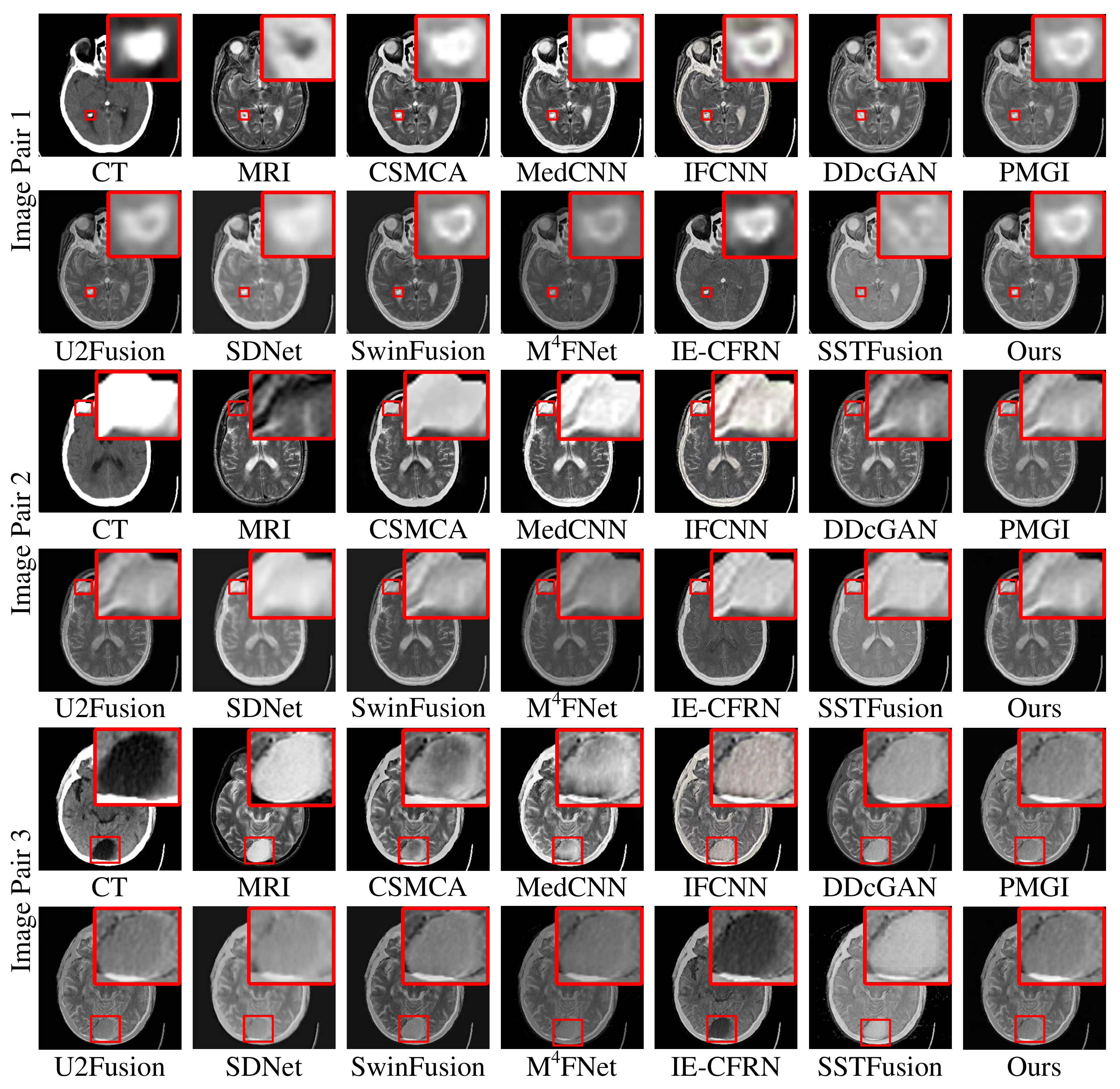}}
	\caption{Qualitative comparisons of proposed AdaFuse with eleven state-of-the-art methods of three typical image pairs on CT-MRI fusion dataset. For a better comparison, a small region is enlarge in the red box.}
	\label{fig_CT_results}
\end{figure*}

Fig. \ref{fig_CT_results} shows the fusion results of three pairs of CT and MRI images obtained by different methods, with a red box indicating the magnified view of the region with significant differences. It can be observed that the fusion results of U2Fusion, SDNet and SSTFusion are overly blurred, where the gray matter regions of the CT images obscure the details of the MRI images, resulting in unclear edges in the fused images. Although the IFCNN model preserves the image details, it exhibits visual artifacts such as color distortion and over-enhancement. For the CSMCA, MedCNN and IE-CFRN methods, the high-density regions in the CT images overshadow the details of the MRI images, leading to the loss of MRI information (highlighted in red boxes in the first and second rows). Regarding the low-density regions in the CT images (indicated by the red box in the third row), after fusion with the corresponding high-brightness MRI images, the DDCGAN, PMGI, SwinFusion and M$^4$FNet produce fusion results that attenuate the texture and detail information in the low-density regions of CT. In comparison to these methods, the proposed AdaFuse can preserve more detailed information from the source images in all scenarios and is not affected by abnormal regions in the CT or MRI images.

\begin{figure*}[!t]
	\centering
	\includegraphics[width=\textwidth]{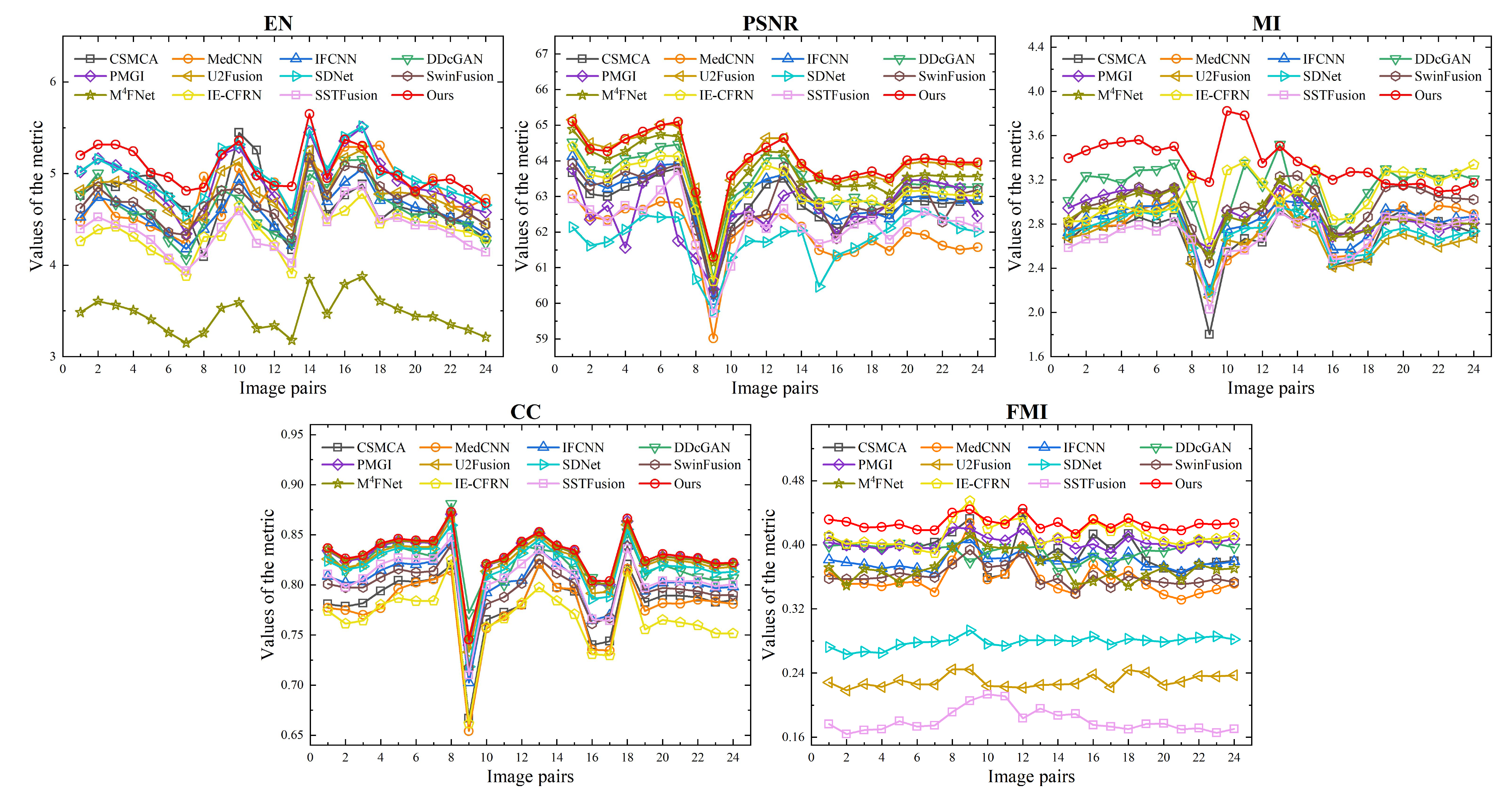}
	\caption{Quantitative comparison results of proposed AdaFuse with eleven state-of-the-art alternatives on 24 image pairs from CT-MRI fusion dataset.}
	\label{fig_CT_results_metric}
\end{figure*}

To quantitatively analyze the differences between different methods, Table \ref{tab_CT_results} provides the quantitative evaluation metrics of different methods on the CT-MRI dataset. It can be observed that our method achieves the highest values for all metrics. In addition to the mean values of the evaluation metrics, we also notice that our method has small standard deviation (STD) for all metrics, indicating better stability and consistent performance in producing good fusion results for test sets. To further demonstrate this advantage, Fig. \ref{fig_CT_results_metric} shows the fusion metric curves for each test image pair. It can be seen that our method (red lines) outperforms other methods in all metrics for most samples. Specifically, for EN metric, our method is optimal on 1-7, 13-14 and 21-24 image pairs. For MI metric, image pairs 1-18 achieves the best results. For PSNR, CC and FMI, our method perform best on nearly all pairs of images. 
\begin{table*}[!t]
	\centering
	\caption{ Mean and standard deviation of five metrics obtained by proposed AdaFuse with eleven state-of-the-art alternatives on 24 test image pairs from CT-MRI fusion dataset. \textcolor{red}{\textbf{Red}} indicates the best result and \textcolor[rgb]{ 0,  .69,  .941}{\textbf{Blue}} indicates the second best result.}
	\resizebox{0.95\linewidth}{!}{
		\begin{tabular}{l|l|lllll}
			\toprule
			Methods &Year& EN & PSNR & MI    & CC    & FMI \\
			\midrule
			\midrule
			CSMCA &2017& 4.700$\pm$0.33 & 62.792$\pm$0.71 & 2.692$\pm$0.24 & 0.783$\pm$0.03 & 0.393$\pm$0.02 \\
			MedCNN &2019& 4.774$\pm$0.30 & 61.949$\pm$0.80 & 2.776$\pm$0.21 & 0.778$\pm$0.03 & 0.351$\pm$0.02 \\
			IFCNN &2020& 4.605$\pm$0.24 & 62.951$\pm$0.80 & 2.817$\pm$0.18 & 0.803$\pm$0.03 & 0.380$\pm$0.01 \\
			DDcGAN &2020& 4.614$\pm$0.28 & 63.378$\pm$0.79 & \textcolor[rgb]{ 0,  .69,  .941}{\textbf{3.132$\pm$0.22}} & 0.821$\pm$0.02 & 0.393$\pm$0.01 \\
			PMGI  &2020& 4.927$\pm$0.30 & 62.611$\pm$0.81 & 2.893$\pm$0.16 & 0.828$\pm$0.02 & 0.403$\pm$0.01 \\
			U2Fusion &2020& 4.792$\pm$0.24 & \textcolor[rgb]{ 0,  .69,  .941}{\textbf{63.976$\pm$0.80}} & 2.679$\pm$0.20 & 0.824$\pm$0.03 & 0.230$\pm$0.01 \\
			SDNet &2021& \textcolor[rgb]{ 0,  .69,  .941}{\textbf{4.986$\pm$0.26}} & 61.795$\pm$0.68 & 2.725$\pm$0.18 & 0.819$\pm$0.03 & 0.279$\pm$0.01 \\
			SwinFusion &2022& 4.685$\pm$0.22 & 62.888$\pm$0.82 & 2.979$\pm$0.18 & 0.797$\pm$0.02 & 0.361$\pm$0.01 \\
			M$^4$FNet &2023&3.459$\pm$0.19 &63.687$\pm$0.77 &2.880$\pm$0.16 &\textcolor[rgb]{ 0,  .69,  .941}{\textbf{0.829$\pm$0.02}} &0.372$\pm$0.02\\
			IE-CFRN &2023&4.375$\pm$0.23&63.196$\pm$0.74&3.068$\pm$0.20&0.767$\pm$0.03 &\textcolor[rgb]{ 0,  .69,  .941}{\textbf{0.413$\pm$0.02}}\\
			SSTFusion &2023&4.384$\pm$0.24 &62.215$\pm$0.75 &2.694$\pm$0.19 &0.806$\pm$0.03 &0.181$\pm$0.01\\
			\midrule
			AdaFuse(Ours)  && \textcolor[rgb]{ 1,  0,  0}{\textbf{5.059$\pm$0.23}} & \textcolor[rgb]{1,0,0}{\textbf{64.001$\pm$0.77}} & \textcolor[rgb]{ 1,  0,  0}{\textbf{3.357$\pm$0.19}} & \textcolor[rgb]{ 1,  0,  0}{\textbf{0.831$\pm$0.02}} & \textcolor[rgb]{ 1,  0,  0}{\textbf{0.427$\pm$0.01}} \\
			\bottomrule
		\end{tabular}%
	}
	\label{tab_CT_results}%
\end{table*}%

\subsubsection{PET-MRI Image Fusion}
\begin{figure*}[!t]
	\centering
	\includegraphics[width=\textwidth]{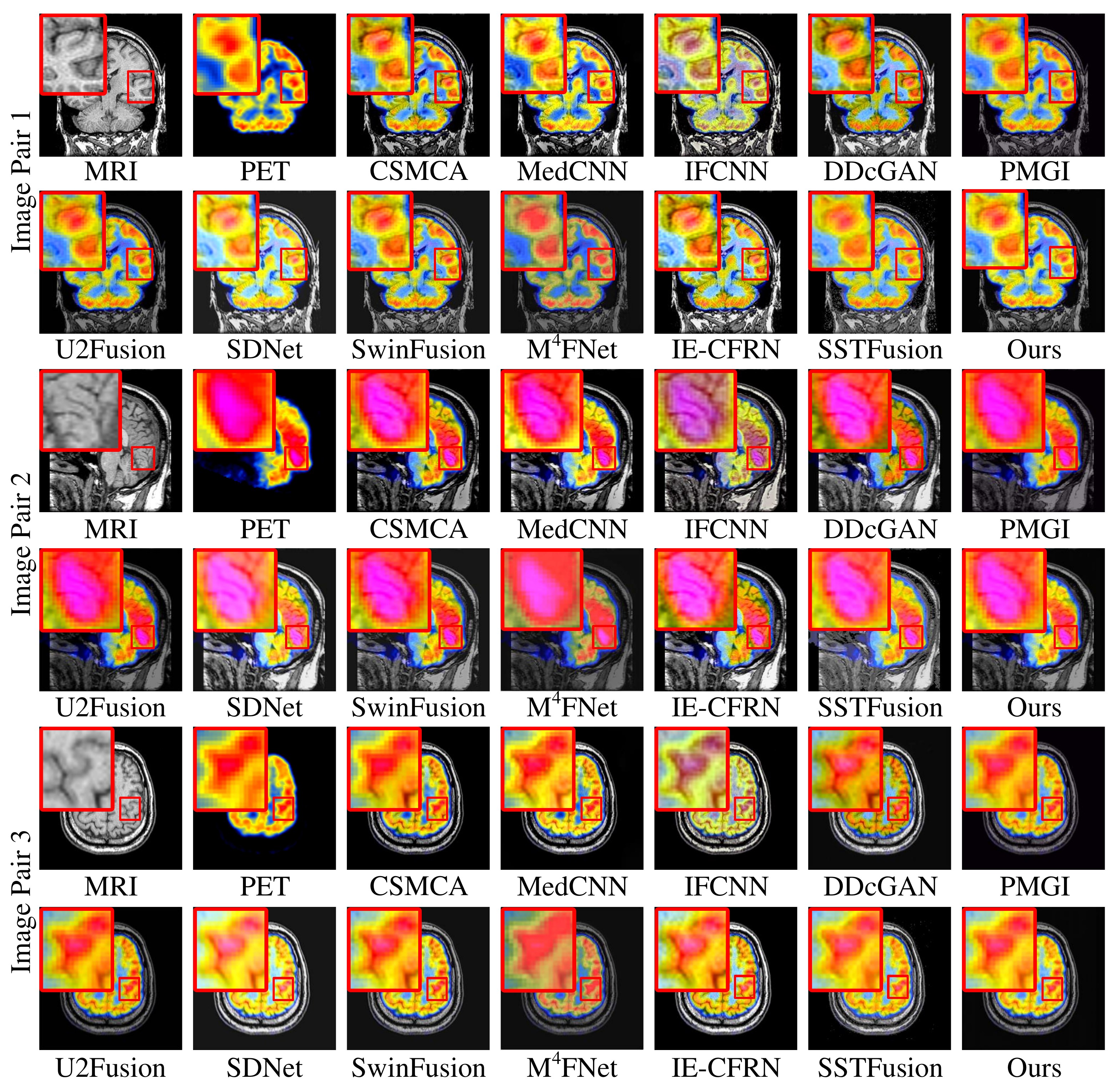}
	\caption{Qualitative comparisons of proposed AdaFuse with eleven state-of-the-art methods of three typical image pairs on PET-MRI fusion dataset. For a better comparison, a small region is enlarge in the red box.}
	\label{fig_PET_results}
\end{figure*}

Fig. \ref{fig_PET_results} shows the fusion results of our method on three pairs of PET-MRI images. The regions with significant differences are zoomed in and highlighted with red boxes. From the preserved PET color information in the fusion results, it can be observed that methods like MedCNN, IFCNN, and SDNet suffer from color distortion, as they only retain the texture part of the MRI image while losing the chromatic information from the PET functional image. In the cases of PMGI, U2Fusion, M$^4$FNet and SSTFusion methods, the grayscale part of the MRI source image is covered by chromatic information, resulting in visual blurring and information loss from the MRI image. Additionally, although DDcGAN, SwinFusion and IE-CFRN methods partially retain important information from both PET and MRI, they generate smooth artifacts at the edges, which affect the quality of the fusion results. In contrast, our proposed AdaFuse preserves the key texture details of MRI image, maintains clear edges under the PET color overlay, and retains important color information of PET image, avoids image distortion, and improves the visual quality of the fusion results.

\begin{figure*}[!t]
	\centering
	\includegraphics[width=\textwidth]{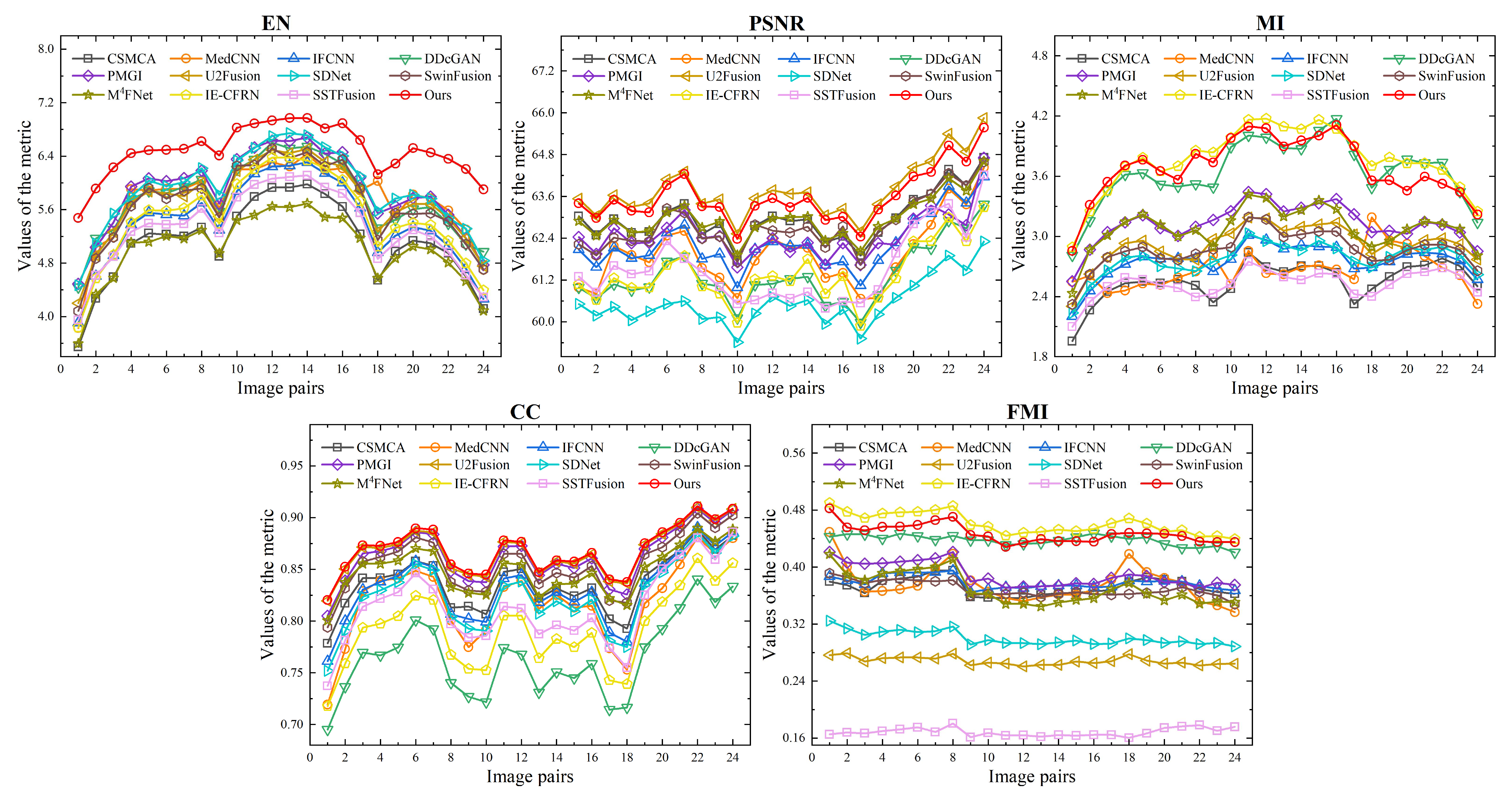}
	\caption{Quantitative comparison results of proposed AdaFuse with eleven state-of-the-art alternatives on 24 image pairs from PET-MRI fusion dataset.}
	\label{fig_PET_results_metric}
\end{figure*}

Table \ref{tab_PET_results} presents the mean and STD of the quantitative evaluation results of our method on the PET-MRI fusion dataset under five evaluation metrics. It can be observed that our method achieves the best values in terms of EN and CC, achieves the suboptimal values in PSNR, MI, CC and FMI, indicating its superior performance compared to other methods. The quantitative evaluation results for each pair of test data are shown in Fig. \ref{fig_PET_results_metric}. It can be visually observed that our method outperforms other methods in terms of EN and CC on all test image pairs. In addition, PSNR, MI and FMI achieve the suboptimal values, only slightly below the best. While IE-CFRN performs well on MI and FMI, it is significantly lower than other methods on other metrics. Overall, our method performs better than all comparison methods for PET-MRI fusion.
\begin{table*}[!t]
	\centering
	\caption{ Mean and standard deviation of five metrics obtained by proposed AdaFuse with eleven state-of-the-art alternatives on 24 test image pairs from PET-MRI fusion dataset. \textcolor{red}{\textbf{Red}} indicates the best result and \textcolor[rgb]{ 0,  .69,  .941}{\textbf{Blue}} indicates the second best result.}
	\resizebox{0.95\linewidth}{!}{
		\begin{tabular}{l|l|lllll}
			\toprule
			Methods &Year& EN & PSNR & MI  & CC  & FMI \\
			\midrule
			\midrule
			CSMCA &2019& 5.112$\pm$0.60 & 62.970$\pm$0.68 & 2.551$\pm$0.19 & 0.837$\pm$0.03 & 0.372$\pm$0.01 \\
			MedCNN &2017& 5.738$\pm$0.58 & 61.950$\pm$0.96 & 2.664$\pm$0.18 & 0.819$\pm$0.04 & 0.377$\pm$0.03 \\
			IFCNN &2020& 5.409$\pm$0.63 & 62.261$\pm$0.77 & 2.752$\pm$0.17 & 0.831$\pm$0.03 & 0.378$\pm$0.01 \\
			DDcGAN &2020& 5.761$\pm$0.55 & 61.382$\pm$0.83 & 3.639$\pm$0.31 & 0.765$\pm$0.04 & 0.438$\pm$0.01 \\
			PMGI  &2020& 5.873$\pm$0.58 & 62.481$\pm$0.65 & 3.125$\pm$0.19 & 0.863$\pm$0.03 & 0.389$\pm$0.02 \\
			U2Fusion &2020& 5.713$\pm$0.57 & \textcolor[rgb]{ 1,  0,  0}{\textbf{63.782$\pm$0.79}} & 2.920$\pm$0.19 & \textcolor[rgb]{ 0,  .69,  .941}{\textbf{0.867$\pm$0.02}} & 0.268$\pm$0.01 \\
			SDNet &2021& \textcolor[rgb]{ 0,  .69,  .941}{\textbf{5.898$\pm$0.58}} & 60.548$\pm$0.67 & 2.771$\pm$0.16 & 0.825$\pm$0.03 & 0.300$\pm$0.01 \\
			SwinFusion &2022& 5.672$\pm$0.60 & 62.782$\pm$0.74 & 2.863$\pm$0.18 & 0.855$\pm$0.03 & 0.369$\pm$0.01 \\
			M$^4$FNet &2023&5.013$\pm$0.51 &62.963$\pm$0.60 &3.078$\pm$0.21 &0.848$\pm$0.02& 0.370$\pm$0.02\\
			IE-CFRN &2023&5.479$\pm$0.65&61.371$\pm$0.82&\textcolor[rgb]{ 1,  0,  0}{\textbf{3.754$\pm$0.31}}&0.792$\pm$0.04 &\textcolor[rgb]{1,0,0}{\textbf{0.461$\pm$0.01}}\\
			SSTFusion &2023&5.311$\pm$0.57 &61.543$\pm$1.01 &2.527$\pm$0.14 &0.814$\pm$0.04 &0.168$\pm$0.01\\
			\midrule
			AdaFuse(Ours) && \textcolor[rgb]{ 1,  0,  0}{\textbf{6.474$\pm$0.37}} & \textcolor[rgb]{ 0,  .69,  .941}{\textbf{63.585$\pm$0.74}} & \textcolor[rgb]{ 0,  .69,  .941}{\textbf{3.681$\pm$0.30}} & \textcolor[rgb]{1,0,0}{\textbf{0.869$\pm$0.02}} & \textcolor[rgb]{0,  .69,  .941}{\textbf{0.447$\pm$0.01}} \\
			\bottomrule
		\end{tabular}%
	}
	\label{tab_PET_results}%
\end{table*}%

\subsubsection{SPECT-MRI Image Fusion}
\begin{figure*}[!t]
	\centering
	\includegraphics[width=\textwidth]{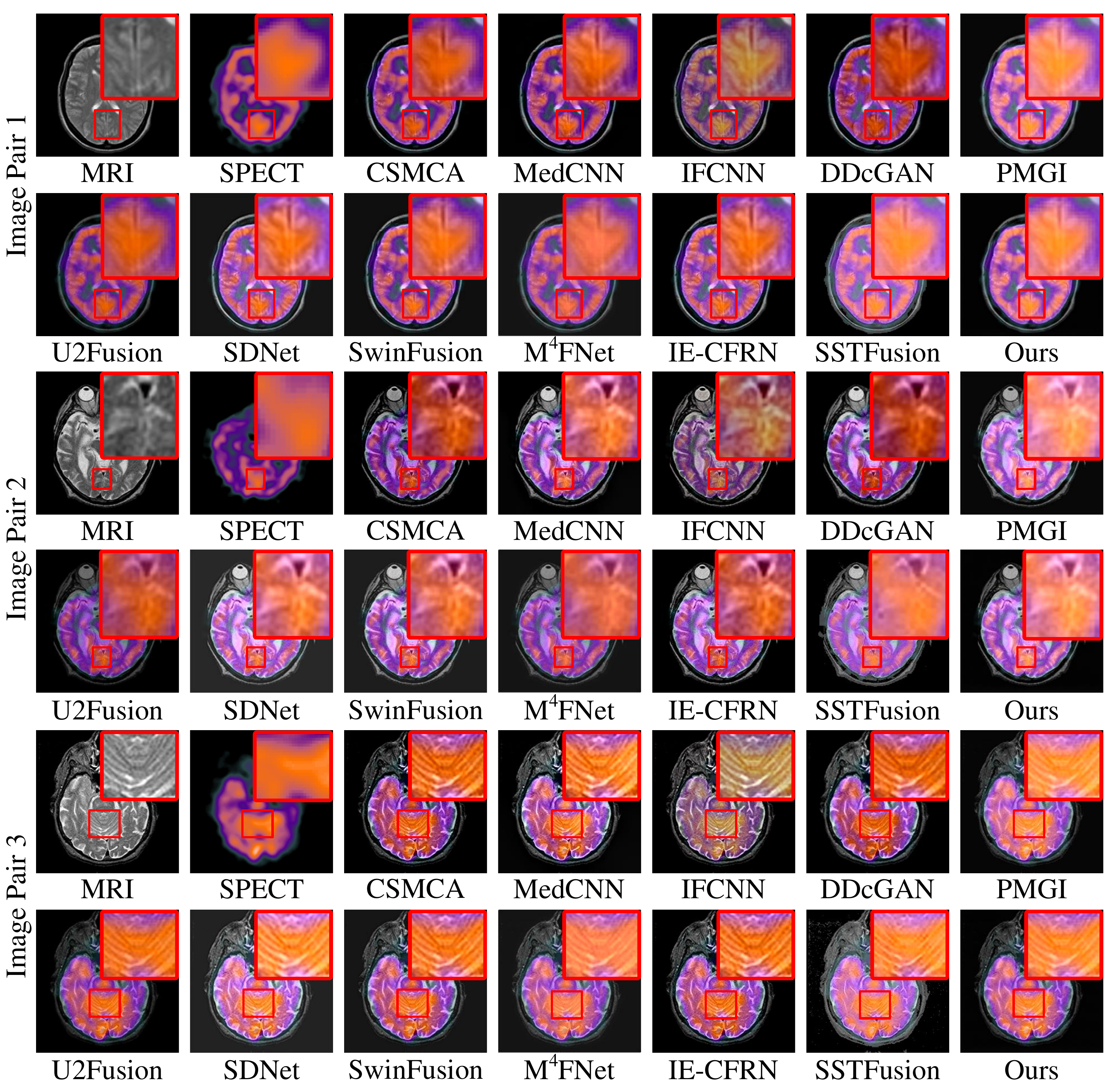}
	\caption{Qualitative comparisons of proposed AdaFuse with eleven state-of-the-art methods of three typical image pairs on SPECT-MRI fusion dataset. For a better comparison, a small region is enlarge in the red box.}
	\label{fig_SPECT_results}
\end{figure*}

Fig. \ref{fig_SPECT_results} shows the fusion results of our method on three pairs of SPECT-MRI images. The visual results indicate that methods such as IFCNN, DDcGAN and PMGI suffer from color distortion. In the fusion results obtained by U2Fusion, SwinFusion and M$^4$FNet, the MRI information appears visually darker, which affects the perception of MRI texture details, while MedCNN, IE-CFRN contains redundant information of MRI. Furthermore, CSMCA, SwinFusion and SSTFusion produce undesirable artifacts that result in smooth and blurry edges. Compared with these methods, our AdaFuse preserves the functional information of the SPECT image, avoids image distortion, and maintains better visualization of MRI texture and edge information. It effectively prevents the loss of critical structural information from the source images.

\begin{figure*}[!t]
	\centering
	\includegraphics[width=\textwidth]{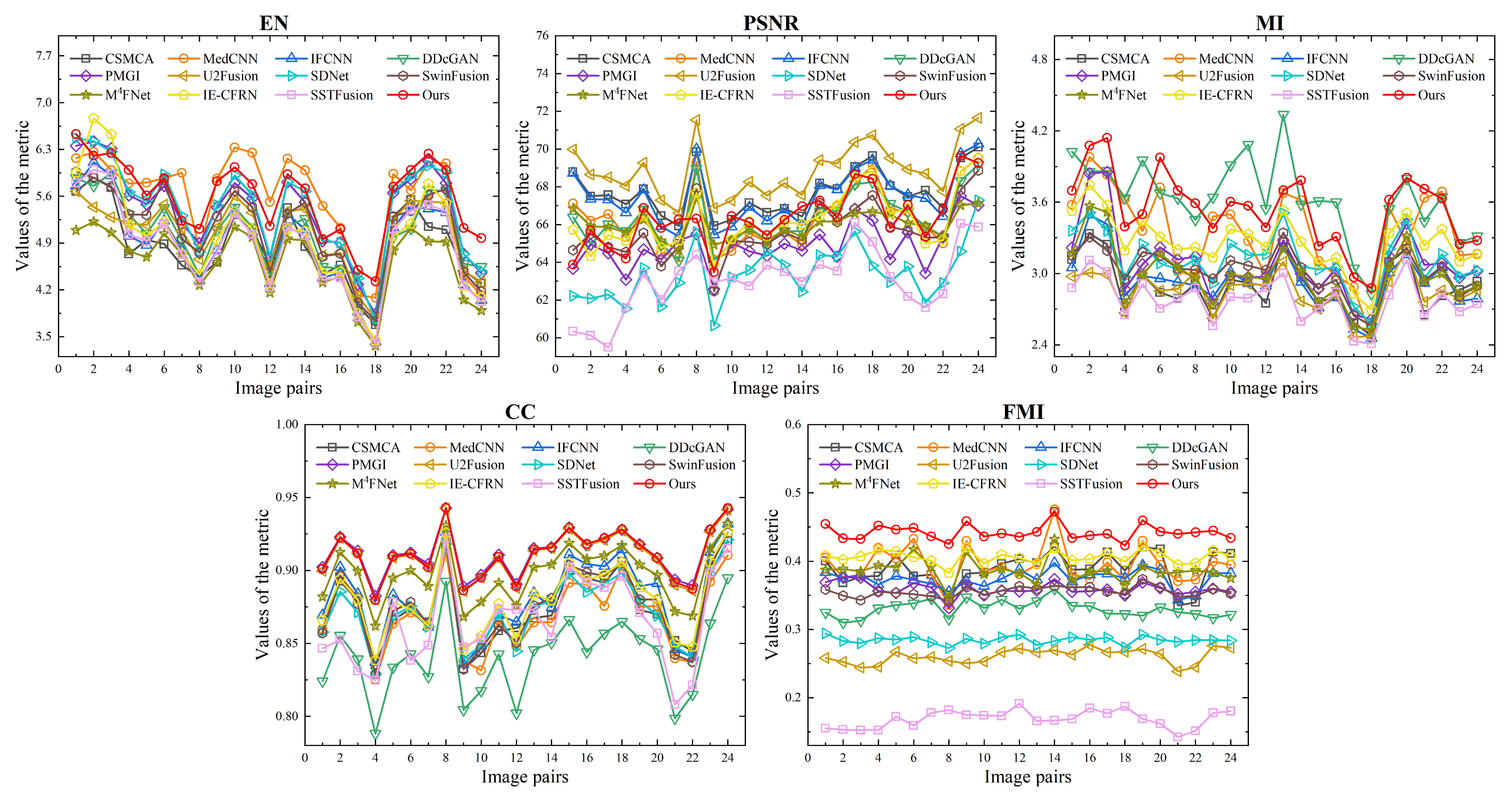}
	\caption{Quantitative comparison results of proposed AdaFuse with eleven state-of-the-art alternatives on 24 image pairs from SPECT-MRI fusion dataset.}
	\label{fig_SPECT_results_metric}
\end{figure*}

Table \ref{tab_SPECT_results} shows the mean and STD of the quantitative evaluation results of our model for the SPECT-MRI fusion dataset under five evaluation metrics. As can be seen from the table, our method achieves two best values as well as two second best values, which are better than the others. Specifically, the best value was achieved for EN, FMI, and the second best for both MI and CC, with a small difference from the best value of 2.41\% and 0.11\%, respectively. On the other hand, PSNR ranked 4th among all methods, with a difference of 3.6\% from the best value. Fig. \ref{fig_SPECT_results_metric} shows the results of the quantitative evaluation for each pair of data. It can be seen that our AdaFuse outperforms most methods in five evaluation metrics. Specifically, our method outperforms the other methods for every pair of data in CC and FMI. For EN, our method is second only to MedCNN overall and achieves the best values for image pairs 16-18, 20-21 and 23-24. In MI, our AdaFuse is second only to DDcGAN and performs more stable, achieving the best values in image pairs 2-3, 6-8, 18-21. However, for PSNR, it ranks in the middle among all compared methods, but shows high stability with less variance. Overall, our method can outperform existing methods in terms of objective evaluation metrics.
\begin{table*}[!t]
	\centering
	\caption{ Mean and standard deviation of five metrics obtained by proposed AdaFuse with eleven state-of-the-art alternatives on 24 test image pairs from SPECT-MRI fusion dataset. \textcolor{red}{\textbf{Red}} indicates the best result and \textcolor[rgb]{ 0,  .69,  .941}{\textbf{Blue}} indicates the second best result.}
	\resizebox{0.95\linewidth}{!}{
		\begin{tabular}{l|l|lllll}
			\toprule
			Methods &Year& EN & PSNR & MI & CC & FMI \\
			\midrule
			\midrule
			CSMCA &2019& 4.884$\pm$0.55 & \textcolor[rgb]{ 0,  .69,  .941}{\textbf{67.708$\pm$1.22}} & 2.898$\pm$0.22 & 0.874$\pm$0.02 & 0.389$\pm$0.02 \\
			MedCNN &2017& \textcolor[rgb]{ 0,  .69,  .941}{\textbf{5.578$\pm$0.72}} & 66.409$\pm$1.60 & 3.354$\pm$0.36 & 0.870$\pm$0.02 & 0.395$\pm$0.03 \\
			IFCNN &2020& 4.909$\pm$0.65 & 67.505$\pm$1.41 & 2.938$\pm$0.23 & 0.881$\pm$0.03 & 0.375$\pm$0.01 \\
			DDcGAN &2020& 5.043$\pm$0.55 & 66.313$\pm$1.40 & \textcolor[rgb]{ 1,  0,  0}{\textbf{3.637$\pm$0.32}} & 0.840$\pm$0.03 & 0.330$\pm$0.01 \\
			PMGI  &2020& 5.391$\pm$0.67 & 64.907$\pm$1.17 & 3.091$\pm$0.29 & \textcolor[rgb]{ 1,  0,  0}{\textbf{0.912$\pm$0.02}} & 0.359$\pm$0.01 \\
			U2Fusion &2020& 5.056$\pm$0.51 & \textcolor[rgb]{ 1,  0,  0}{\textbf{68.875$\pm$1.42}} & 2.852$\pm$0.17 & 0.910$\pm$0.02 & 0.261$\pm$0.01 \\
			SDNet &2021& 5.431$\pm$0.69 & 63.411$\pm$1.48 & 3.121$\pm$0.22 & 0.873$\pm$0.02 & 0.285$\pm$0.01 \\
			SwinFusion &2022& 5.140$\pm$0.61 & 65.692$\pm$1.39 & 3.025$\pm$0.18 & 0.876$\pm$0.03 & 0.356$\pm$0.01 \\
			M$^4$FNet &2023&4.631$\pm$0.49 &65.974$\pm$0.71 &2.990$\pm$0.25 &0.897$\pm$0.02& 0.386$\pm$0.02\\
			IE-CFRN &2023&4.987$\pm$0.77&66.261$\pm$1.48&3.272$\pm$0.23&0.880$\pm$0.02 &\textcolor[rgb]{ 0,  .69,  .941}{\textbf{0.406$\pm$0.01}}\\
			SSTFusion &2023&4.875$\pm$0.66 &63.082$\pm$1.70 &2.782$\pm$0.18 &0.865$\pm$0.03 &0.169$\pm$0.01\\
			\midrule
			AdaFuse(Ours) & & \textcolor[rgb]{ 1,  0,  0}{\textbf{5.592$\pm$0.57}} & 66.383$\pm$1.52 & \textcolor[rgb]{ 0,  .69,  .941}{\textbf{3.549$\pm$0.31}} & \textcolor[rgb]{ 0,  .69,  .941}{\textbf{0.911$\pm$0.02}} & \textcolor[rgb]{ 1,  0,  0}{\textbf{0.442$\pm$0.01}} \\
			\bottomrule
		\end{tabular}%
	}
	\label{tab_SPECT_results}%
\end{table*}%

\subsection{Ablation Study}
We perform ablation studies of our proposed method in this subsection, including the cross attention fusion (CAF) block, the Fourier transform-guided fusion branch (FGFB) in SFF, and the loss function we designed. Without loss of generality, all our ablation studies are performed on the CT-MRI fusion dataset.
\subsubsection{Effectiveness of CAF}
\begin{figure}[!t]
	\centering
	\includegraphics[width=0.55\columnwidth]{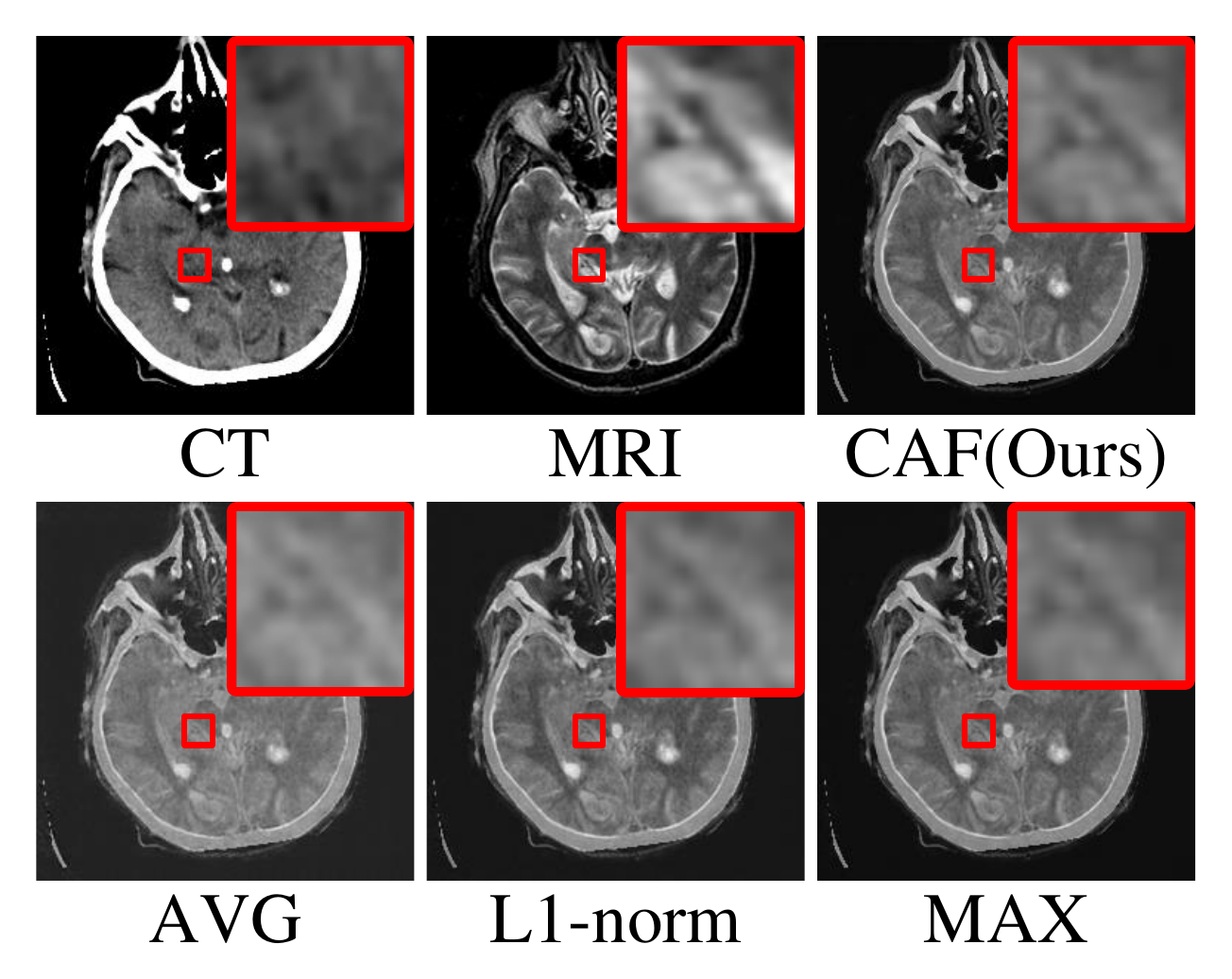}
	\caption{Quantitative comparison of different fusion strategy.}
	\label{fig_ab_CAF}
\end{figure}

To verify that our proposed CAF can effectively adaptively fuse features without human intervention, we replace the cross attention fusion module, introduce other hand-designed fusion rules in the model, including the averaging strategy, the L1-Norm fusion rule \cite{li2018densefuse} and the maximum selection rule \cite{li2020laplacian}, and then compared with our proposed method. The quantitative results of the ablation experiments in Table \ref{tab_ab_CAF} shows that the introduction of CAF block can significantly improve the quality of fusion results. Specifically, AdaFuse achieves the better results in PSNR, MI, CC, and FMI metrics than those using other fusion strategies, with MI and FMI being 15.09\% and 15.68\% higher, respectively. Although the introduction of the CAF leads to a decrease in EN, it improves the fusion performance overall. We show the visualization results in Fig. \ref{fig_ab_CAF}. The average fusion strategy approach leads to smooth edges and blurred texture details, and the L1-norm and maximum fusion strategies do not yield clear fusion results and produced artifacts, although the information in MRI was preserved to some extent. In contrast, our proposed CAF fusion strategy obtains clearer images.
\begin{table*}[!t]
	\centering
	\caption{ Quantitative comparison results of ablation studies of CAF. \textcolor{red}{\textbf{Red}} indicates the best result.}
	\resizebox{0.6\linewidth}{!}{
		\begin{tabular}{llllll}
			\toprule
			Fusion Strategy & EN & PSNR & MI & CC & FMI \\
			\midrule
			AVG   & 5.4081  & 63.1804  & 2.7841  & 0.8267  & 0.3496  \\
			L1-norm & \textcolor[rgb]{ 1,  0,  0}{\textbf{5.6207 }} & 63.7856  & 2.7581  & 0.8265  & 0.3597  \\
			MAX & 5.2555  & 63.4460  & 2.8503  & 0.8288  & 0.3582  \\
			\textbf{CAF(Ours)} & 5.0592  & \textcolor[rgb]{ 1,  0,  0}{\textbf{64.0004 }} & \textcolor[rgb]{ 1,  0,  0}{\textbf{3.3571 }} & \textcolor[rgb]{ 1,  0,  0}{\textbf{0.8306 }} & \textcolor[rgb]{ 1,  0,  0}{\textbf{0.4266 }} \\
			\bottomrule
		\end{tabular}%
	}
	\label{tab_ab_CAF}%
\end{table*}%

\subsubsection{Effectiveness of FGFB}
\begin{figure}[!t]
	\centering
	\includegraphics[width=0.8\columnwidth]{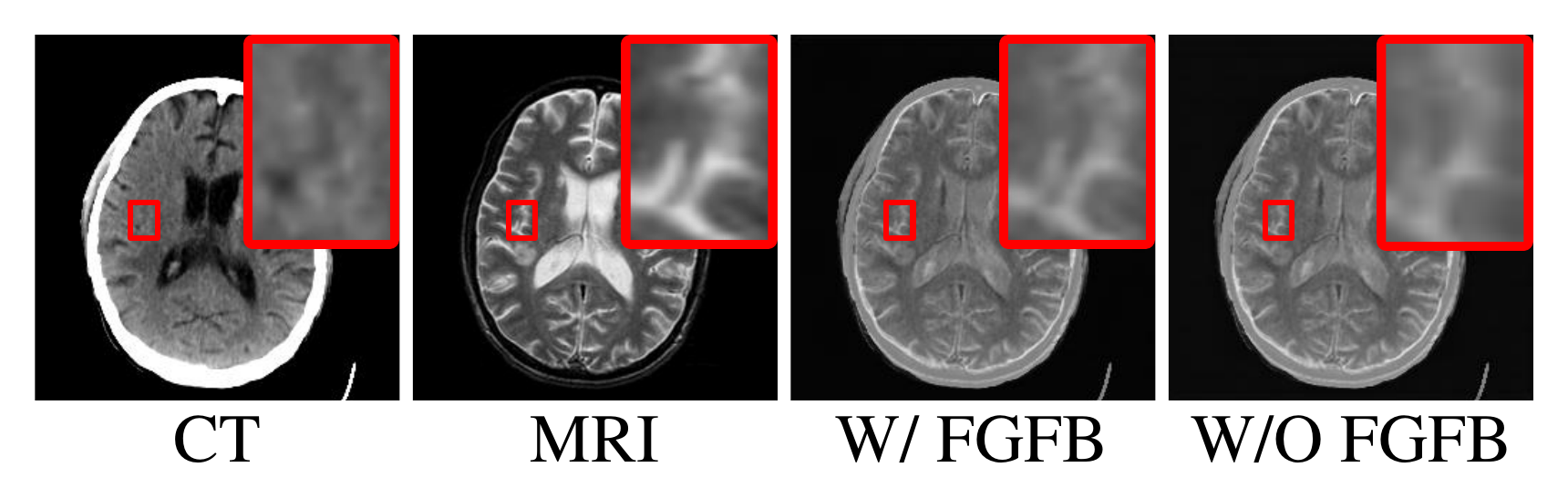}
	\caption{Quantitative evaluation to verify the effectiveness of FGFB.}
	\label{fig_ab_FGFB}
\end{figure}

To compensate for the shortage of the Transformer model to extract high-frequency features, we introduce the Fourier transform-guided fusion branch FGFB in the fusion network, which uses frequency-domain features to guide adaptive global feature fusion. To verify the effectiveness of our proposed Fourier transform-guided branch, we remove the Fourier transform-guided branch and then compare it with our method. The results of the experimental quantitative evaluation are shown in Table \ref{tab_ab_FTGB}. It can be seen that our method is improved in four evaluation metrics, PSNR, MI, CC, and FMI, after introducing FGFB. From the results visualized shown in Fig. \ref{fig_ab_FGFB}, we find that the high-frequency texture information is clearer in the fusion results obtained after the introduction of FGFB, while only the low-frequency information in the images is retained after the removal of FGFB, and the edge details of MRI are lost.
\begin{table*}[!t]
	\centering
	\caption{ Quantitative comparison results of ablation studies of FTGB. \textcolor{red}{\textbf{Red}} indicates the best result.}
	\resizebox{0.6\linewidth}{!}{
		\begin{tabular}{clllll}
			\toprule
			Method  & EN & PSNR & MI & CC & FMI \\
			\midrule
			W/O FGFB & \textcolor[rgb]{ 1,  0,  0}{\textbf{5.4651 }} & 63.6365  & 3.1420  & 0.8300  & 0.4249  \\
			W/ FGFB & 5.0592  & \textcolor[rgb]{ 1,  0,  0}{\textbf{64.0004 }} & \textcolor[rgb]{ 1,  0,  0}{\textbf{3.3571 }} & \textcolor[rgb]{ 1,  0,  0}{\textbf{0.8306 }} & \textcolor[rgb]{ 1,  0,  0}{\textbf{0.4266 }} \\
			\bottomrule
		\end{tabular}%
	}
	\label{tab_ab_FTGB}%
\end{table*}%

\subsubsection{Effectiveness of Loss Function}
\begin{figure}[!t]
	\centering
	\includegraphics[width=\columnwidth]{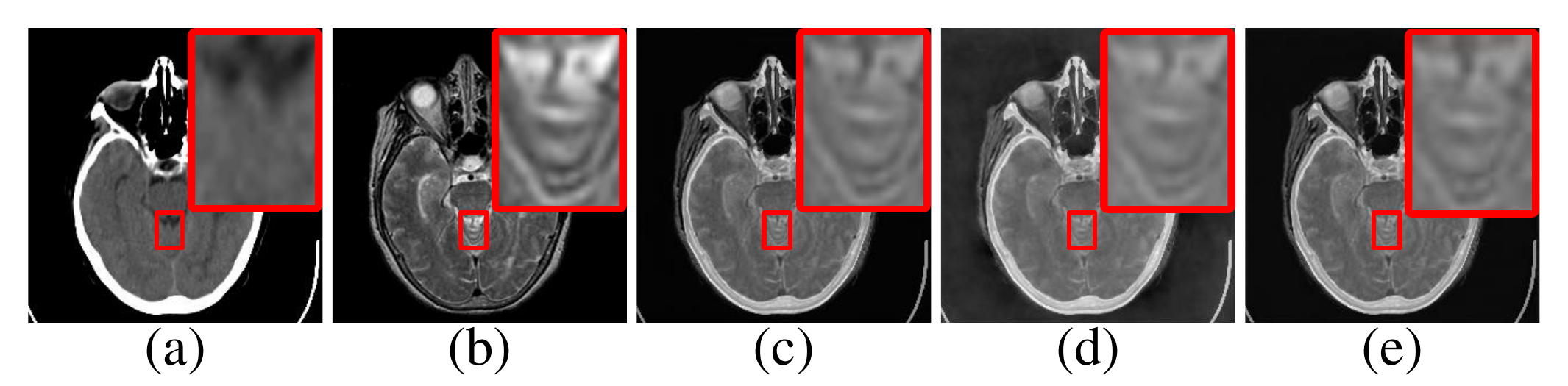}
	\caption{Quantitative evaluation to verify the effectiveness of loss function, where (a)CT image, (b)MRI image, (c)with $\mathcal{L}_{content}$ and $\mathcal{L}_{structure}$, (d)only with $\mathcal{L}_{structure}$, (e)only with $\mathcal{L}_{content}$.}
	\label{fig_ab_loss}
\end{figure}

In this work, we design a new loss function in our proposed method to better obtain high-quality fusion results. To verify the effectiveness of the loss function, we remove the content term and the structure term of the loss function separately, and then compare them with our adopted loss function. Quantitative results are shown in Table \ref{tab_ab_loss}. When only the structure term ${\mathcal{L}_{structure}}$  or the content term ${\mathcal{L}_{content}}$  of the loss function is retained, the EN is improved but the remaining four evaluation metrics are all significantly lower, and the optimal results can be obtained by using the loss function we designed. Further, the qualitative results in Fig. \ref{fig_ab_loss} show that when only the structure items are retained, the fusion results produce a large number of artifacts, and the texture structure is too smooth. When the loss function has only the content items retained, the edges are blurred. In contrast, training with our loss function can yield higher quality fusion results.
\begin{table*}[!t]
	\centering
	\caption{ Quantitative comparison results of ablation studies of Loss Function. \textcolor{red}{\textbf{Red}} indicates the best result.}
	\resizebox{0.65\linewidth}{!}{
		\begin{tabular}{cc|lllll}
			\toprule
			$\mathcal{L}_{content}$ & $\mathcal{L}_{structure}$ & EN & PSNR & MI & CC & FMI \\
			\midrule
			& \ding{52} & \textcolor[rgb]{ 1,  0,  0}{\textbf{6.6268 }} & 62.3632  & 2.6306  & 0.7974  & 0.4078  \\
			\ding{52}     &       & 5.4425  & 63.8337  & 3.1293  & 0.8298  & 0.4239  \\
			\ding{52}     & \ding{52} & 5.0592  & \textcolor[rgb]{ 1,  0,  0}{\textbf{64.0004 }} & \textcolor[rgb]{ 1,  0,  0}{\textbf{3.3571 }} & \textcolor[rgb]{ 1,  0,  0}{\textbf{0.8306 }} & \textcolor[rgb]{ 1,  0,  0}{\textbf{0.4266 }} \\
			\bottomrule
		\end{tabular}%
	}
	\label{tab_ab_loss}%
\end{table*}%

\section{Discussion}
In this work, we propose AdaFuse, a deep learning model for multi-modal medical image fusion. We compare our method with state-of-the-art fusion methods on three multi-modal medical image fusion datasets: CT-MRI, PET-MRI, and SPECT-MRI. The results demonstrate that our approach facilitates feature adaptive fusion and preserves high-frequency information, leading to superior performance compared to existing methods.

It is well known that Transformer models have strong capabilities in learning global contextual features, which makes them outperform traditional and CNN-based methods. Therefore, we incorporate Transformers for global contextual feature extraction at each scale. However, Transformers tend to prioritize global features, specifically low-frequency information, while disregarding crucial high-frequency texture details present in medical images. Existing approaches often combine convolutional neural networks with Transformers for feature extraction or design loss functions for deep constraints, but these methods do not fundamentally address the limitations of Transformers. Benefiting from the powerful ability of Fourier transform to extract high-frequency features, we propose a dual-branch fusion approach called spatial-frequential fusion (SFF) strategy that introduces a Fourier-guided fusion branch. This branch leverages frequency domain features obtained through Fourier transform to guide cross-domain fusion in the spatial domain, thereby compensating for the high-frequency features extracted by Transformers, as shown in the lower part of Fig. \ref{fig_Framework}. The ablation results in Fig. \ref{fig_ab_FGFB} and Table. \ref{tab_ab_FTGB} demonstrate that our proposed method effectively preserves the texture information and enhances edge clarity. Furthermore, for the frequency domain features, we design a logarithmic loss function based on the structure tensor to constrain the frequency domain features, which avoids the issue of imbalanced training between the two branches and enables better utilization of frequency features. The ablation experiments in Fig. \ref{fig_ab_loss} and Table \ref{tab_ab_loss} also validate the effectiveness of the loss function.

In addition, many existing studies cannot adaptively compute the feature weights of source images and fail to better preserve critical information. In comparison methods, CSMCA, MedCNN, IFCNN, M$^4$Fnet and SSTFusion adopt traditional fusion strategies, resulting in artifacts, while DDCGAN, PMGI, U2Fusion and SDNet concat source images as a whole, which will lead to information loss of source images. On the contrary, SwinFuison and IE-CFRN obtained better results by calculating the fusion feature weights of source images. In our method, we propose cross attention fusion (CAF) block (as shown in Fig. \ref{fig_FusionNetwork}). The multi-head self-attention mechanism enables the dynamic computation of adaptive weights for each patch block, thereby preserving more important information in the fusion result. By utilizing the CAF, we can fuse the features of the two source images, adaptively obtaining high-quality fusion results without the need for complex manual design. The ablation results (as shown in Fig. \ref{fig_ab_CAF} and Table \ref{tab_ab_CAF}) indicate that our proposed method better preserves the texture information in the source images, producing fusion results without introducing additional noise compared to traditional methods.

All comparative experimental results demonstrate that our proposed AdaFuse outperforms existing fusion methods and does not require complex manual design for fusion strategies. The optimal results of the four quantitative evaluation metrics, namely EN, MI, CC, and FMI, objectively demonstrate the outstanding performance of our method in preserving the information from different modalities of the source images. The good PSNR reflects the structural integrity and avoids image distortion issues. From the visual results, it is evident that our method avoids problems such as excessive contrast, image distortion, and blurring, which also indicates the ability of our proposed method to adaptively fuse image features. The clearer edges further confirm the effectiveness of our spatial-frequency cross-domain fusion method that incorporates frequency domain features. It is worth noting that in the ablation experiment results, there is an increase in the EN value. This can be attributed to the removal of the specific module we designed, which resulted in undesirable artifacts or the retention of redundant features. Consequently, the fused image contains additional irrelevant information, leading to an increase in information quantity but potentially compromising the quality of the fusion results.

While we have demonstrated the effectiveness of our method in the experiments, there are still limitations to our proposed approach. Firstly, our method stacks Transformer Encoders at four scales, and although it has been shown that Transformers have superior performance, this may impose limitations on the running speed of our method. Therefore, it is necessary for us to consider strategies to reduce FLOPS while maintaining performance. Secondly, although fusing the frequency domain features through Fourier transform improves fusion performance in the spatial domain, it is important to note that we did not operate in the complex domain after the Fourier transform. Instead, we converted it back to the spatial domain for fusion, which may result in the loss of crucial complex domain features. Therefore, future research could explore methods that incorporate complex neural networks. Additionally, due to the limited  availability of labeled data for unsupervised medical image fusion, many current approaches have adopted self-supervised training for fusion models and have achieved good performance \cite{zhao2021self,liang2022fusion,zhang2023multimodal}. Therefore, our future work will focus on improving the proposed method in terms of performance, efficiency, and training strategies.

\section{Conclusion}
In this paper, we propose a spatial-frequential fusion network named AdaFuse to address the challenges of high-frequency information retaining and adaptive fusion rules designing in the field of multi-modal medical image fusion. To adaptively fuse the low and high frequential information of multi-modal images, we combine frequency-domain features extracted by Fourier transform, and then fuse the multi-modal features in spatial and frequential domains respectively through the cross-attention module by exchanging the query and key vectors of different modalities. Moreover, to retain the details of the fused image, we calculate the cross attention between the spatial and frequential fusion features to further guide the spatial-frequential information fusion. In addition, we design a novel loss function to preserve more complementary information by combining content loss and structure loss. Through the comparison and ablation experiments on several publicly available datasets, we have demonstrated that the proposed method outperforms state-of-the-art methods, and the proposed frequential-guided cross attention fusion strategy and multi-losses are indeed beneficial for medical image fusion. 









\section*{Acknowledgement}
This work was partially funded by the National Natural Science Foundations of China (Grant No.62161004), Guizhou Provincial Science and Technology Projects (QianKeHe ZK [2021] Key 002), and Guizhou Provincial Science and Technology Projects (QianKeHe ZK [2022] 046).


\bibliographystyle{cas-model2-names}
\bibliography{references.bib}

%
\end{sloppypar}
\end{document}